%% file: main.tex
\documentclass{article} 
\usepackage{iclr2027_conference,times}
\usepackage{multirow}
\usepackage{colortbl}
\usepackage{wrapfig,lipsum,booktabs}
\usepackage{xcolor}
\usepackage{breqn}
\usepackage{skeldoc}
\usepackage{adjustbox} 
\usepackage{caption}
\input{math_commands.tex}

\usepackage{cleveref}

\usepackage{hyperref}
\usepackage{url}
\usepackage{changes}

\title{Representation World Model:\\Learning States, Transition and Executable Plans in Representation}

\iclrfinalcopy

\author{\textbf{Yijun Yuan, Weicheng Zheng, Weibang Wang, Minghui Qin, Chang Sun,} \\\textbf{Junhao Huang, Kenan Li, Anmin Liu, Yicheng Yao, Hang Zhao}\\
IIIS, Tsinghua University\\
\color{blue}{\tt\small{\url{https://tsinghua-mars-lab.github.io/RepresentationWorldModel}}}\\
{\{yuanyj, hangzhao\}@mail.tsinghua.edu.cn}\\
}
\begin{document}

\maketitle

\begin{center}
    \includegraphics[width=1.0\linewidth]{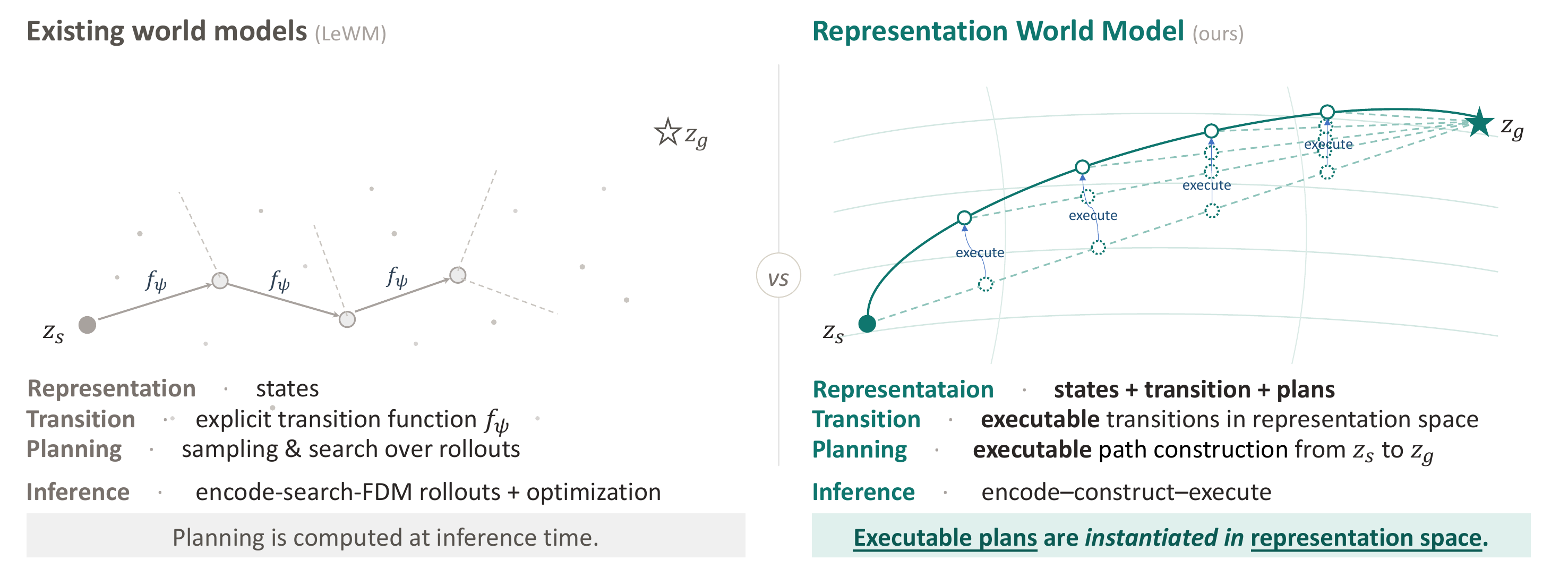}
    \captionof{figure}{
    \textbf{Representing states, transitions, and executable plans in latent space.}
    Existing world models such as LeWM represent states in latent space, while modeling transitions with an explicit dynamics function $f_\psi$, and planning through rollout-based sampling and search at inference time (left). 
RWM instead learns a latent geometry that jointly represents states, transitions, and executable plans, turning otherwise unconstrained regions of the representation space into actionable paths connecting $z_s$ and $z_g$ (right).
}
    \label{fig:teaser}
\end{center}

\begin{abstract}
We propose the \emph{Representation World Model} (RWM), which learns states, transitions, and executable plans directly in representation space.
Unlike existing world models that typically learn latent representations together with explicit dynamics models and perform planning through search, optimization, or policy-based prediction, RWM directly incorporates planning into the learned representation geometry.
RWM learns the representation geometry by applying inverse-dynamics supervision locally along latent paths constructed from endpoint representations, requiring these paths to preserve task-relevant state and transition information.
At inference, planning is performed by directly constructing a latent path between the current and goal representations, with inverse dynamics used to recover the corresponding actions, without recursive rollouts or action-space search.
Experiments on continuous-control benchmarks demonstrate the effectiveness of RWM for direct planning, while results on robotic manipulation further show its potential to extend to more complex embodied control tasks.
These results suggest that planning directly in representation space provides a promising alternative to conventional world-model planning.
\end{abstract}

\input{tex/intro_v3}

\input{tex/rela_v3}

\input{tex/method_v3}
\input{tex/exp_v2}
\input{tex/conclusion}

\bibliography{iclr2027_conference}
\bibliographystyle{iclr2027_conference}

\appendix
\input{tex/appendix}

\end{document}

%% file: math_commands.tex
\usepackage{amsmath,amsfonts,bm}

\def\eqref#1{equation~\ref{#1}}

\def\1{\bm{1}}

\DeclareMathAlphabet{\mathsfit}{\encodingdefault}{\sfdefault}{m}{sl}
\SetMathAlphabet{\mathsfit}{bold}{\encodingdefault}{\sfdefault}{bx}{n}



%% file: tex/intro_v3.tex
\section{Introduction}
Recent advances in foundation models have demonstrated the power of scalable representation learning across vision, language, and robotics~\citep{radford2021clip,oquab2024dinov2,brown2020gpt3,
driess2023palme,zitkovich2023rt2}. 
A good representation should preserve and organize the information needed by downstream tasks.
For embodied agents, such relevance is naturally defined by action and interaction. Visual--action trajectories therefore provide a direct source for learning action-relevant representations, motivating recent latent world models that incorporate actions into representation learning~\citep{lewm,gao2026fastlewm}.

However, existing approaches still rely on proxy prediction together with predefined representation priors such as SIGReg. Such priors are introduced to prevent representation collapse under forward prediction, but impose fixed assumptions on the latent space that may become restrictive as data and tasks scale~\citep{boylan2026no}. 
Meanwhile, proxy objectives encourage representations to preserve whatever is useful for prediction, including state information that may be irrelevant to action~\citep{liu2026temporally}. 
Moreover, they do not directly require the latent geometry itself to be organized around task-relevant transitions or planning~\citep{li2026predictive}. 
These limitations motivate learning representations more directly from action-level supervision, where the task signal itself rules out the trivial collapsed solution and determines both what information should be emphasized and how it should be structured.

Yet direct action supervision introduces a second challenge: when the encoder and planner are jointly learned, the action loss constrains only their combined output, leaving the division between representation and planning underdetermined. A strong planner may compensate for weak representation structure, while an overly simple planner leaves planning computations that the encoder cannot absorb, making end-to-end training ineffective.

We therefore ask: \textbf{How can action-level supervision be directed into the representation itself?}

Our key idea is to constrain planning to a direct geometric construction in representation space and apply action supervision to the resulting latent transitions, thereby directing the learning signal toward the representation geometry itself. Based on this idea, we introduce the \emph{Representation World Model} (RWM), which learns task-relevant states, transitions, and executable plans directly in a shared representation space, as illustrated in Fig.~\ref{fig:teaser}.

Given an action-labeled trajectory, RWM constructs a latent path between its endpoint representations and applies a shared inverse dynamics model locally along the path to recover the corresponding actions. This objective directly shapes the representation to preserve action-relevant information and organize it into a geometry that supports executable plan construction.

At inference time, RWM encodes the current and goal observations,
directly constructs a latent path between them, and decodes actions locally along the path.
Planning is therefore realized through geometric path construction
followed by action decoding, without recursive forward rollouts
or action-sequence search.
The same construction used during training provides the latent plan, while inverse dynamics translates its local transitions into actions.
We evaluate this formulation on continuous-control and robotic
manipulation benchmarks, demonstrating its utility for search-free control.

The contributions of this paper are as follows:
\begin{itemize}
\item We introduce the \emph{Representation World Model} (RWM), a formulation that learns action-relevant states, transitions, and planning structure within a shared representation space.

\item We propose an inverse dynamics objective on constructed latent paths, which directly uses action supervision to shape both the information preserved by the representation and the geometry in which executable plans can be constructed.

\item We demonstrate better search-free control performance on continuous-control benchmarks and further show the potential of RWM for more complex embodied control tasks.
\end{itemize}



%% file: tex/rela_v3.tex
\section{Related Work}

\subsection{Representation Learning}

Representation learning seeks to map high-dimensional observations into latent variables that preserve information useful for downstream tasks. Early approaches based on autoencoders and variational autoencoders learned representations primarily through reconstruction, while later self-supervised methods shifted the focus toward invariance and semantic structure. 
This line of work, including
InstDisc~\citep{wu2018unsupervised}, MoCo~\citep{he2020momentum}, SimCLR~\citep{chen2020simple}, BYOL~\citep{grill2020bootstrap}, Barlow Twins~\citep{zbontar2021barlow}, VICReg~\citep{bardes2022vicreg}, DINO~\citep{caron2021emerging}, and iBOT~\citep{zhou2022ibot}, showed that transferable visual representations can emerge from large-scale self-supervision without explicitly reconstructing observations.

More recent work has incorporated temporal structure into representation learning. 
Predictive approaches such as V-JEPA~\citep{bardes2024vjepa} learn by predicting latent representations across time, while embodied variants~\citep{lewm,nam2026cjepa} further model state transitions and interaction dynamics in latent space for continuous-control.
Related approaches exploit temporal structure and reachability through planning-aligned objectives to shape latent representations for downstream decision making~\citep{li2026predictive}. 
This moves learned state representations beyond static visual structure toward transition- and decision-relevant structure.

RWM takes a further step by structuring not only represented states, but also the latent regions between them. A fixed path-construction rule provides a simple geometric scaffold, while inverse-dynamics supervision shapes the representation onto this scaffold, turning constructed paths into executable plans.

\subsection{World Models}

World models learn internal representations of an environment together with mechanisms for modeling its evolution. 
Early latent world models~\citep{ha2018world} established the use of compact latent variables for sequential prediction, while PlaNet~\citep{hafner2019planet} and the Dreamer family~\cite{hafner2020dreamer,hafner2021dreamerv2,hafner2025dreamerv3} demonstrated that learned latent dynamics can support effective model-based control.
MuZero~\citep{schrittwieser2020muzero} further showed that representations and learned dynamics can be optimized specifically for planning without reconstructing the underlying observations.

As world models scale to richer visual environments, generative systems such as Genie~\citep{bruce2024genie} and Cosmos~\citep{nvidia2025cosmos} model future evolution by explicitly generating action-conditioned visual observations. 
While expressive, such visual imagination can be computationally expensive for downstream control, as each candidate future requires generating high-dimensional observations. 
This motivates recent latent world models that predict directly in compact representation spaces. 
LeWM~\citep{maes2026lewm}, for example, learns action-conditioned latent dynamics from visual--action trajectories without reconstructing future observations, enabling substantially more efficient latent-space planning.

However, such latent representations are still primarily shaped by predictive objectives. Avoiding collapse requires additional latent-space regularization such as SIGReg, introducing fixed assumptions that may become restrictive as data and task increase. Besides, prediction may also preserve behavior-irrelevant factors and fail to learn a geometry aligned with task-relevant search and planning. RWM takes a further step by grounding representation learning directly in action supervision, allowing both state information and planning structure to be learned within the representation space.

\subsection{Planning with Learned World Models}
A common planning strategy of using latent world model is to imagine and evaluate future trajectories at inference time. Methods such as CEM or MPPI sample candidate action sequences, predict their outcomes, and select actions according to task objectives. Recent latent approaches, including DINO-WM~\citep{zhou2025dinowm}, LeWM~\citep{lewm}, and Fast-LeWM~\citep{gao2026fastlewm}, improve the underlying representations and predictive models, but planning still relies on evaluating candidate futures at inference time.

Instead, another approach amortizes goal-directed control into a learned policy or inverse dynamics model. Goal-conditioned policies directly predict actions from the current state and target, avoiding iterative online search. GC-IDM~\citep{nguyen2026gcidm} follows this paradigm by recovering actions from current and goal representations.

RWM provides a third planning paradigm. Unlike search-based methods, it does not evaluate candidate futures; unlike goal-conditioned policies, it does not directly map endpoints to actions. Instead, RWM constructs a latent path between the current and goal states and decodes the path into actions, making planning a direct path-construction process in representation space.

%% file: tex/method_v3.tex
\section{Method}
\label{sec:method}
Our goal is to learn a representation 
{in which states, actionable transitions, and the structure required for executable planning are jointly expressed in the latent geometry}. We first formalize the problem, and then introduce how RWM constructs and learns such a representation space.

\subsection{Problem Setup}
\label{sec:problem_setup}

We consider goal-conditioned control from an offline dataset $\mathcal D$ of action-labeled trajectories
$\tau=(o_0,a_0,o_1,a_1,\ldots,a_{T-1},o_T)$,
where $o_t\in\mathcal O$ denotes an observation (or fixed-length observation chunk) and $a_t\in\mathcal A$ denotes the action (or fixed-length action chunk), that connects two consecutive observations.
An encoder $f_\theta:\mathcal O\rightarrow\mathcal Z$ maps each observation to a latent representation
\begin{equation}
z_t=f_\theta(o_t) \in\mathcal Z.
\end{equation}

Conventional latent world models additionally learn a forward transition model
$f_\psi:\mathcal Z\times\mathcal A\rightarrow\mathcal Z$,
such that
$z_{t+1}=f_\psi(z_t,a_t)$.
Given start and goal states $(z_s,z_g)$, planning then requires searching over candidate actions through repeated rollouts:
\begin{equation*}
\mathbf a^*
=
\arg\min_{\mathbf a=(a_s,\ldots,a_{s+H-1})}
d\left(f_\psi^{(H)}(z_s,\mathbf a),z_g\right),
\end{equation*}
where $f_\psi^{(H)}$ denotes an $H$-step rollout and $d$ is a latent discrepancy.

\begin{figure}[t]
\centering
\includegraphics[width=\linewidth]{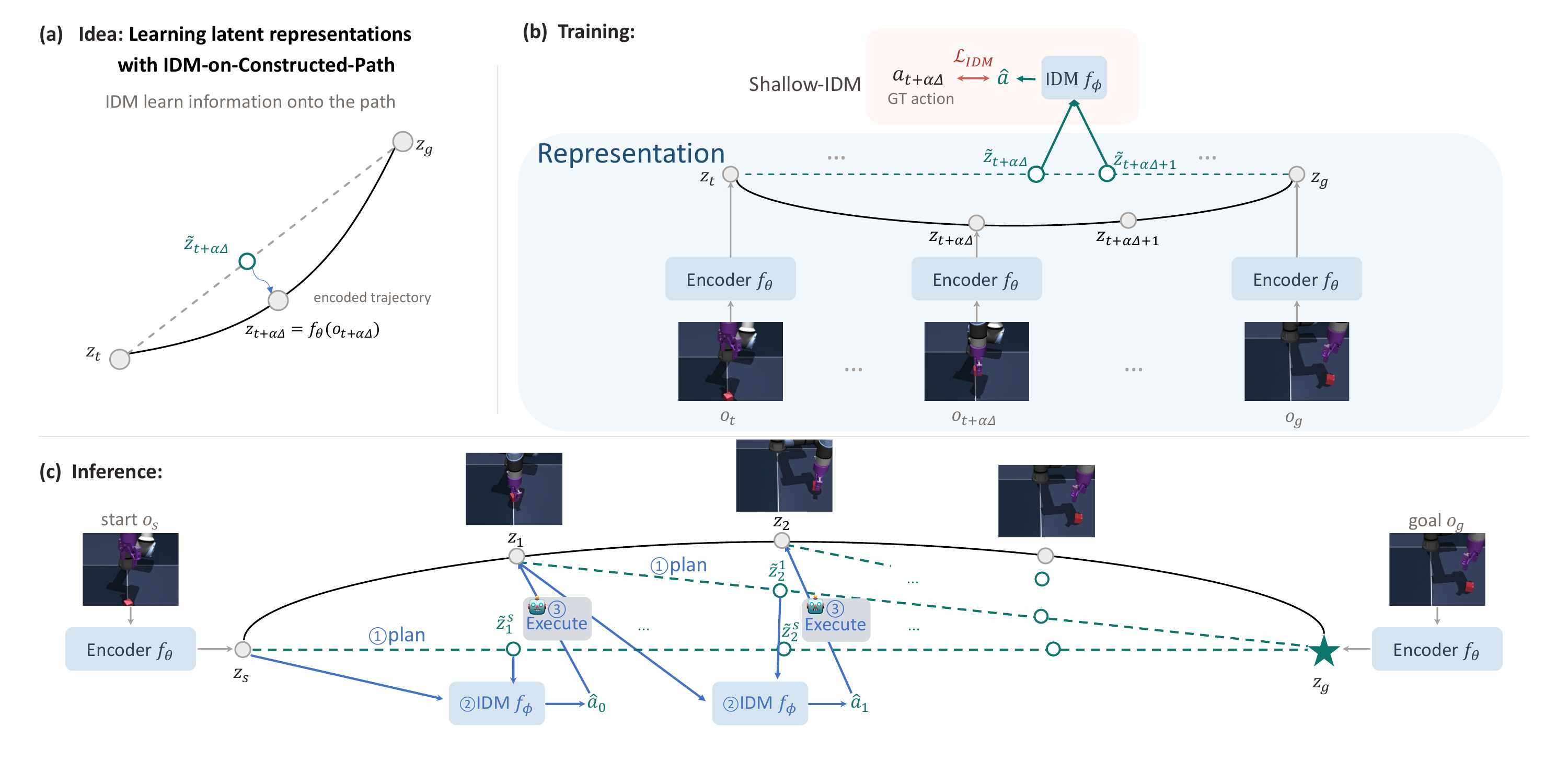}
\vspace{-1cm}
\caption{
\textbf{Overview of RWM}.
During training, RWM constructs {latent paths} between encoded states and {learns their executable structure through action supervision}. At inference, it constructs paths between current and goal, and decodes locally on it, enabling closed-loop control.
}
\label{fig:method}
\end{figure}

\subsection{Representation World Model}
\label{sec:rwm}
Instead of relying on recursive forward prediction or search, RWM learns actionable transition structure along constructed latent paths. Planning can then be realized by directly constructing a path between the current and goal states and decoding its transitions into actions, as illustrated in Fig.~\ref{fig:method}.

Consider two observations $o_t$ and $o_u$ from the same trajectory, where $t<u$.
After encoding the endpoints as
$z_t=f_\theta(o_t)$ and $z_u=f_\theta(o_u)$,
we construct the representation at temporal position $m$, where $t<m<u$, by
\begin{equation*}
    \tilde z_m
=
(1-\alpha_{t,m,u})z_t
+
\alpha_{t,m,u}z_u,
\quad
\alpha_{t,m,u}
=
\frac{m-t}{u-t}.
\end{equation*}

{Importantly, $\tilde z_m$ is only a geometrically constructed latent intermediate before learning; interpolation alone does not imply a valid transition or executable trajectory. RWM therefore trains the representation so that these constructed intermediates acquire task-relevant state, transition, and action semantics.}

\emph{This construction exposes the previously unconstrained regions between observed states to learning.} Because every intermediate point is determined by the endpoint representations, supervision applied along the constructed path also shapes the endpoints and, consequently, the geometry of the representation space itself. The objectives introduced in Sec.~\ref{sec:training} make this geometry informative about the actions required to traverse them.

\paragraph{Planning as a geometric operation in representation space.}

The same construction can be used directly for planning.
Given current and goal observations $(o_t,o_g)$, RWM first encodes
$z_t=f_\theta(o_t)$,
$z_g=f_\theta(o_g)$,
and constructs intermediate transition representations
$\tilde z^{t}_m
=
(1-\alpha_{t,m,g})z_t+\alpha_{t,m,g} z_g$,
with $\alpha_m=\frac{m-t}{g-t}$,
$m=t+1,\ldots,g-1$.

Together with
$\tilde z^{t}_t=z_t$
and
$\tilde z^{t}_g=z_g$,
these representations define a latent transition path
\begin{equation*}    
\mathcal P_{s:g}
=
(\tilde z^{t}_t, \tilde z^{t}_{t+1}, \ldots,\tilde z^{t}_g).
\end{equation*}

Unlike rollout-based planning, each {intermediate representation} is constructed directly from the encoded endpoints, without recursively predicting intermediate states.
The key challenge is therefore to make the constructed path actionable, such that it can serve as an executable plan. We address this through action-grounded representation learning.

\subsection{Action-Grounded Representation Learning}
\label{sec:training}

For a sampled trajectory segment
$(o_t,a_t,\ldots,a_{u-1},o_u)$
we encode
$z_i=f_\theta(o_i)$,
and construct transition representations along the segment:
$\tilde z_i
=
(1-\alpha_i)z_t+\alpha_i z_u$, with
$\alpha_i=\frac{i-t}{u-t}$,
$i=t,\ldots,u$.

Training is majorly with inverse dynamics that provides action-relevant supervision along the constructed path.

\paragraph{Action supervision on constructed path.}
Interpolation alone only specifies latent locations between two endpoints.
There is no guarantee that the resulting path represents how the system actually transitions between states.
We therefore train an inverse dynamics model
$f_\phi:\mathcal Z\times\mathcal Z\rightarrow\mathcal A$
to recover the action between neighboring interpolated representations:
\begin{equation}
\mathcal L_{\mathrm{IDM}}^{\mathrm{int}}=
\frac{1}{u-t}
\sum_{i=t}^{u-1}
\left|
f_\phi(\tilde z_i,\tilde z_{i+1})-a_i
\right|_2^2.
\label{eq:idm_int}
\end{equation}

This objective grounds the constructed path in action semantics: neighboring interpolated representations must support recovery of the actions executed along the demonstrated trajectory. 
It therefore encourages the regions between encoded states to represent actionable transitions rather than arbitrary geometric intermediates.

Optionally, we additionally apply inverse dynamics directly to neighboring encoded states,
$\mathcal L_{\mathrm{IDM}}^{\mathrm{enc}}=
\frac{1}{u-t}
\sum_{i=t}^{u-1}
\left|
f_\phi(z_i,z_{i+1})-a_i
\right|_2^2.
$
which provides a direct action-grounding signal on observed latent transitions.
This term is not essential to the construction, but can serve as an auxiliary supervision for the encoded representation.

\paragraph{Why IDM on interpolation shapes the representation.}
IDM first prevents trivial feature collapse: if all observations shared the same representation, identical latent inputs would correspond to different actions, making the IDM objective unable to satisfy.

It also encourages the encoder to preserve transition-relevant state. If distinct object configurations collapse to similar latent endpoints, they induce similar interpolated paths, creating conflicting IDM supervision.

Thus, interpolation-based IDM grounds the constructed path in action semantics and encourages the representation to retain the state information required to recover the correct actions along it.

\paragraph{Soft interpolation consistency.}
Optionally, we softly align each constructed transition representation with the encoded state at the corresponding temporal position:
$\mathcal L_{\mathrm{consist}}=
\frac{1}{u-t-1}
\sum_{i=t+1}^{u-1}
\left|
\tilde z_i-z_i
\right|_2^2$.

Unlike IDM, this objective does not determine what information the representation should preserve. It only discourages the constructed path from drifting away from the geometry of observed trajectories.

\paragraph{Overall objective.}
The complete objective is
\begin{equation}
\mathcal L_{\mathrm{all}}
=
\mathcal L_{\mathrm{IDM}}^{\mathrm{int}}
+
\lambda_{\mathrm{IDM}}^{\mathrm{enc}}
\mathcal L_{\mathrm{IDM}}^{\mathrm{enc}}
+
\lambda_{\mathrm{consist}}
\mathcal L_{\mathrm{consist}}.
\label{eq:total_loss}
\end{equation}

The three terms provide complementary constraints on the learned representation.
$\mathcal L_{\mathrm{IDM}}^{\mathrm{int}}$ grounds endpoint-constructed transitions in executable actions and propagates this supervision through the interpolated path to the endpoint representations.
$\mathcal L_{\mathrm{IDM}}^{\mathrm{enc}}$ optionally provides direct action supervision on observed latent transitions.
$\mathcal L_{\mathrm{consist}}$ supplies soft geometric supervision by aligning constructed intermediates with the corresponding observation-induced states.
During training, only the encoder $f_\theta$ and IDM $f_\phi$ are learned; no FDM is required.

\subsection{Rollout-Free Inference}
\label{sec:inference}

At inference, RWM only encodes the current and goal observations.
Following Fig.~\ref{fig:method} (c), it directly constructs a sequence of intermediate transition representations between the two encoded endpoints.
The shared IDM then recovers the action associated with each consecutive pair:
$\hat a_m
=
f_\phi
\left(
\hat z^{t}_{m},
\hat z^{t}_{m+1}
\right)_
{m=t,\ldots,g-1}
$.

Because all intermediate representations are constructed directly from the encoded endpoints, the action sequence can be decoded in parallel for open-loop execution. For closed-loop control, RWM re-encodes the current observation after each action chunk and re-constructs the latent plan toward the goal.

Inference thus follows a direct \emph{encode--construct--decode} procedure: the learned representation geometry turns endpoint interpolation into an executable latent plan, requiring neither recursive forward prediction nor action-space search.

%% file: tex/exp_v2.tex
\section{Experiments}
\label{sec:experiments}

\subsection{Experimental Setup}
\label{sec:exp_setup}

\paragraph{Continuous-control benchmarks.}
Full training and implementation details of RWM are provided in Appendix~\ref{app:training_setup}.
For evaluation, RWM use the same task definitions and evaluation protocol as INTACT variants.
We evaluate PushT, Cube, Reacher, and TwoRoom following INTACT~\cite{intact}. Our comparisons include DINO-WM~\citep{zhou2025dinowm}, LeWM~\citep{maes2026lewm}, Fast-LeWM~\citep{gao2026fastlewm}, Qantara~\citep{rakhimov2026qantara}, PRISM~\citep{wang2026prism}, C-JEPA~\citep{nam2026cjepa}, GC-IDM~\citep{nguyen2026gcidm} and INTACT~\citep{intact}.
Which are classified as search-based and direct methods.
With their correspondingly control type listed in inference column of Table~\ref{tab:continuous-control}.

\paragraph{LIBERO-Goal.}
We further evaluate the ten LIBERO-Goal tasks. The comparison follows the setting in RC-aux~\citep{li2026predictive} for LeWM~\cite{maes2026lewm}, RC-aux~\citep{li2026predictive}, and RWM; OpenVLA-OFT 7B~\citep{kim2025fine} is included as a large pretrained external reference.
Implementation details of RWM specific to the goal-unobserved LIBERO-Goal setting are provided in Appendix~\ref{app:libero_goal}.

We evaluate whether RWM can scale to visually complex multi-task manipulation. We report model size (B) and success rate (SR) for latent world models and one external large pretrained reference.

\subsection{Continuous-Control Results}
\label{sec:continuous_results}

\begin{table*}[t]
\centering
\scriptsize
\setlength{\tabcolsep}{3.0pt}
\renewcommand{\arraystretch}{1.0}
\caption{\textbf{Success rate (\%) on continuous-control benchmarks.}
Published rows are external references and are not paired controls; $\dagger$ denotes partial task coverage or a distinct reproduction protocol from INTACT~\cite{intact}.}
\label{tab:continuous-control}
\begin{adjustbox}{max width=\textwidth}
\begin{tabular}{lllcrrrrr}
\toprule
Class & Method & Inference & PushT & Cube & Reacher & TwoRoom & Macro \\
\midrule
\multirow{10}{*}{Search-based}
& DINO-WM~\citep{zhou2025dinowm} & CEM 
& $74.0\!\pm\!4.5$ & $86.0\!\pm\!4.7$ & $79.0\!\pm\!5.1$ & $100.0\!\pm\!.0$ & 84.75 \\
& LeWM~\citep{maes2026lewm} & CEM $300{\times}(30/10)$ 
& $\mathbf{96.0\!\pm\!4.0}$ & $74.0\!\pm\!3.0$ & $86.0\!\pm\!5.0$ & $87.0\!\pm\!2.5$ & 85.75 \\
& Fast-LeWM~\citep{gao2026fastlewm} & CEM 
& 96.0 & 80.0 & 88.0 & 98.0 & 90.50 \\
& Fast-LeWM + SC~\citep{gao2026fastlewm} & CEM + score 
& 98.0 & 82.0 & 90.0 & 98.0 & 92.00 \\
& Qantara$^\dagger$~\citep{rakhimov2026qantara} & CEM $300{\times}30$, $K=4$ 
& $90.1\!\pm\!1.1$ & $93.7\!\pm\!.7$ & $80.9\!\pm\!1.8$ & $100.0\!\pm\!.0$ & 91.18 \\
& PRISM$^\dagger$~\citep{wang2026prism} & MPPI $128{\times}30$ 
& $89\!\pm\!4$ & $79\!\pm\!6$ & -- & -- & -- \\
& C-JEPA$^\dagger$~\citep{nam2026cjepa} & CEM $300{\times}30$ 
& 88.67 & -- & -- & -- & -- \\

& INTACT~\citep{intact}  & Pure CEM $300{\times}30$ 
& $88.44\!\pm\!1.17$ & $68.44\!\pm\!.77$ & $83.67\!\pm\!.67$ & $82.89\!\pm\!.84$ & $80.86\!\pm\!.51$ \\
& INTACT~\citep{intact}  & Actor-guided CEM $300{\times}30$ 
& $93.56\!\pm\!.96$ & $96.89\!\pm\!.19$ & $86.67\!\pm\!.88$ & $98.00\!\pm\!1.15$ & $93.78\!\pm\!.77$ \\
& INTACT~\citep{intact}  & Guarded actor $128{\times}3$
& $92.22\!\pm\!.69$ & $99.78\!\pm\!.19$ & $97.44\!\pm\!.77$ & $98.00\!\pm\!1.15$ & $96.86\!\pm\!.38$ \\
\midrule
\multirow{3}{*}{Direct}
& GC-IDM~\citep{nguyen2026gcidm} &  Goal-conditioned IDM 
& $84.7\!\pm\!5.0$ & $99.3\!\pm\!1.2$ & $\mathbf{100.0\!\pm\!.0}$ & $\mathbf{100.0\!\pm\!.0}$ & 96.00 \\
& INTACT~\citep{intact} &  Goal-conditioned IDM 
& $85.78\!\pm\!1.54$ & $100.00\!\pm\!.00$ & $97.67\!\pm\!.00$ & $97.89\!\pm\!1.26$ & $95.33\!\pm\!.58$ \\

& \textbf{RWM} (Ours) & Representation plan + IDM 
&  $\mathbf{95.44\!\pm\!0.69}$ & $\mathbf{100.00\!\pm\!.00}$ & $\mathbf{97.89\!\pm\!.19}$ & $99.67\!\pm\!.00$ & $\mathbf{98.25\!\pm\!.14}$ \\
\bottomrule
\end{tabular}
\end{adjustbox}
\end{table*}
We follow the INTACT evaluation protocol: each model trained with one of three training seeds is evaluated over three evaluation seeds, and we report the mean and standard deviation across the three training seeds. As shown in Table~\ref{tab:continuous-control}, RWM achieves $95.44\%$, $100.00\%$, $97.89\%$, and $99.67\%$ success rate on PushT, Cube, Reacher, and TwoRoom, respectively, yielding a macro average of $\mathbf{98.25\%}$. RWM outperforms prior search-based methods while requiring no online trajectory search, including the strongest matched INTACT variant ($96.86\%$ macro SR) that evaluates 384 candidate trajectories at inference time. RWM also exceed over the goal-conditioned IDM INTACT baseline ($95.33\%$), with the largest gain on PushT ($85.78\%\rightarrow95.44\%$). 
These results show that learning executable planing structure into the representation enables high quality search-free control.

\subsection{LIBERO-Goal Results}
\label{sec:libero_results}

As shown in Table~\ref{tab:libero-goal}, RWM achieves an average success rate of $\mathbf{93.0\%}$ on LIBERO-Goal, substantially outperforming the LeWM ($71.2\%$) and RC-aux ($81.2\%$) baselines. 
The improvement is particularly clear on challenging tasks; for example, on T5, RWM reaches $90\%$ success compared with $44\%$ for LeWM and $48\%$ for RC-aux. 
Overall, RWM achieves strong performance across the task suite with a smaller model, demonstrating that the proposed representation-space path construction scales effectively to visually complex robotic manipulation. 
OpenVLA-OFT achieves a higher average SR of $97.0\%$ with a much larger $7$B pretrained model, serving only as an external reference while highlighting the strong potential of RWM.

\begin{table*}[htbp]
\centering
\scriptsize
\setlength{\tabcolsep}{4.0pt}
\renewcommand{\arraystretch}{1.05}
\caption{\textbf{Per-task success rate on LIBERO-Goal.}
}
\label{tab:libero-goal}
\begin{adjustbox}{max width=\textwidth}
\begin{tabular}{lcccccccccccc}
\toprule
Method & Model size& T0 & T1 & T2 & T3 & T4 & T5 & T6 & T7 & T8 & T9 & Mean \\
\midrule
LeWM + OFT-head & 0.07 B
& 0.64 & 0.78 & 0.70 & 0.76 & 0.90 & 0.44 & 0.60 & 0.94 & 0.70 & 0.66 & 0.712 \\
RC-aux + OFT-head& 0.07B
& 0.92 & 0.86 & 0.80 & 0.78 & 0.96 & 0.48 & $\mathbf{0.70}$ & 0.96 & 0.86 & 0.80 & 0.812 \\
\textbf{RWM (Ours)}& $\mathbf{0.03B}$
&$\mathbf{1.00}$&$\mathbf{0.98}$&$\mathbf{0.94}$
&$\mathbf{0.92}$
&$\mathbf{0.96}$
&$\mathbf{0.90}$
&0.64
&$\mathbf{1.00}$
&$\mathbf{1.00}$
&$\mathbf{0.96}$
&$\mathbf{0.930}$ \\
\midrule
OpenVLA-OFT & 7B
& 0.98 & 0.92 & 0.96 & 0.86 & 1.00 & 1.00 & 1.00 & 1.00 & 1.00 & 0.98 & {0.970} \\
\bottomrule
\end{tabular}
\end{adjustbox}
\end{table*}


\subsection{Ablation studies}
Our ablations examine if IDM supervision learn informative representation, how replanning frequency affects closed-loop control, and how each training objective contributes to performance.
Unless otherwise specified, all ablation models are trained with seed $3072$.

\paragraph{Can RWM learn informative representations without explicit representation regularization?}

LeWM learns its representation through forward-dynamics prediction together with explicit representation regularization, whereas RWM {learns its representation through action-grounded and geometric objectives without such explicit representation regularization}.
We therefore ask whether {RWM nevertheless preserves state- and action-relevant information comparable to that captured by LeWM}.

We freeze the trained encoders and train identical tiny MLP probes for one epoch on the same training data.
The state probe predicts the simulator configuration from a single encoded representation $z$, excluding velocities, while the action probe predicts the action from a pair of consecutive representations $(z_t,z_{t+1})$.
We report $R^2$ on the episode-split validation set.
\begin{figure}[htbp!]
    \centering
    \includegraphics[width=\linewidth]{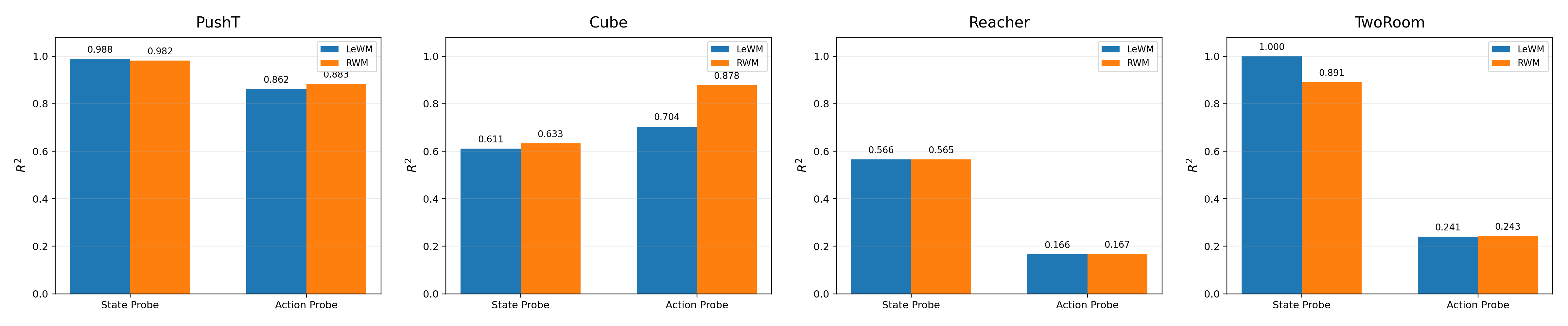}
    \caption{
    \textbf{State and action probing of LeWM and RWM encoder representations.}
    }
    \label{fig:state_probe}
\end{figure}

As shown in Fig.~\ref{fig:state_probe}, RWM achieves broadly comparable state and action recoverability to LeWM across the four environments.
State information is similarly recoverable on PushT, Cube, and Reacher, although RWM is lower on TwoRoom.
For action prediction, RWM matches LeWM on PushT, Reacher, and TwoRoom and substantially improves over it on Cube.
These results suggest that RWM can learn representations that preserve much of the physical-state and action-relevant information captured by LeWM, without relying on forward-dynamics prediction or explicit representation regularization.

\paragraph{Do constructed representations preserve state and action information?}
Having established that the encoded representation $z$ captures meaningful physical information, we next examine whether the constructed representation $\tilde{z}$ preserves the same information when moving off the observed-state manifold.

We freeze the trained RWM encoder and train identical tiny MLP probes for one epoch on the same training data.
For state probing, we train the probe only on encoded representations $z$, and evaluate it on both $z$ and constructed representations $\tilde{z}$, excluding velocities.
For action probing, we train only on observed-state pairs $(z_t,z_{t+1})$ and evaluate action prediction on three pair types:
$(z_t,z_{t+1})$, $(z_t,\tilde{z}_{t+1})$, and $(\tilde{z}_t,\tilde{z}_{t+1})$.
We report $R^2$ on the episode-split validation set.
\begin{figure}[htbp]
    \centering
    \includegraphics[width=\linewidth]{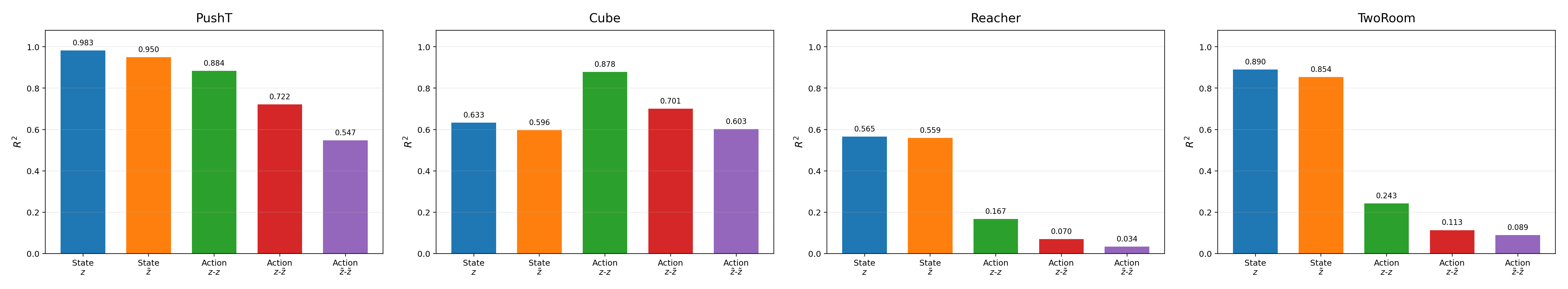
    }
    \caption{
    \textbf{Information retained by interpolated RWM representations.}
    }
    \label{fig:interp_probe}
\end{figure}

As shown in Fig.~\ref{fig:interp_probe}, physical state remains highly recoverable from $\tilde{z}$, with performance close to that of the encoded representation $z$ across all four environments.
Moreover, actions remain recoverable when one or both encoded states are replaced by constructed representations, particularly on PushT and Cube.
Although action recoverability decreases as more constructed representations are involved, these results show that $\tilde{z}$ is not merely an arbitrary point between two latent codes: it retains substantial physical-state information and action-relevant structure required for transition execution. 

\begin{figure*}[htbp!]
    \centering
    \includegraphics[width=\textwidth]{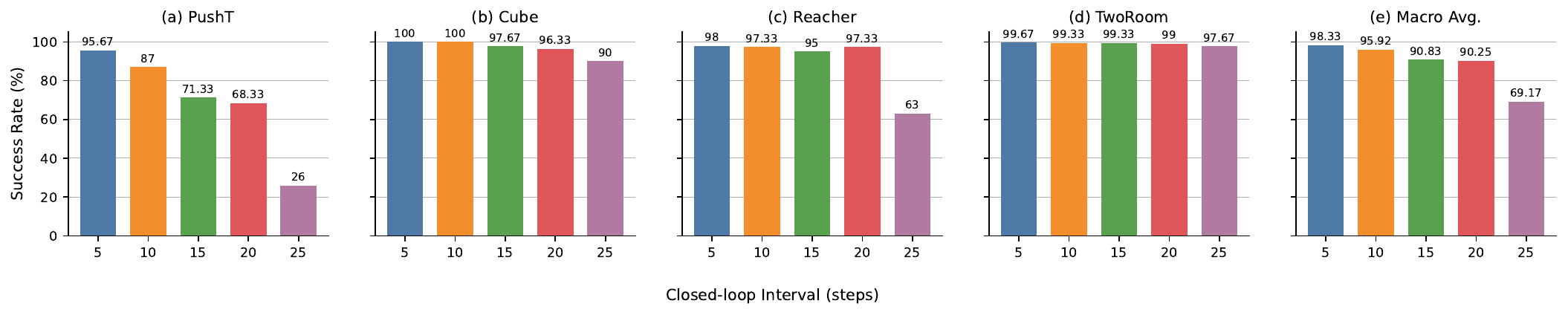}
    \caption{
    \textbf{Effect of the closed-loop interval.}
    }
    \label{fig:closeloop_ablation}
\end{figure*}

\paragraph{Effect of the closeloop interval.}
We vary the number of executed actions between two consecutive replanning steps,
$h\in\{5,10,15,20,25\}$, while keeping the trained model fixed.
As shown in Fig.~\ref{fig:closeloop_ablation}, more frequent replanning generally improves control performance.
RWM achieves the highest macro-average SR of $98.33\%$ at $h=5$, which we use as the default setting.
Performance remains strong for moderate intervals, but degrades as the execution becomes increasingly open-loop, with the macro-average SR dropping to $69.17\%$ at $h=25$.
The degradation is mainly driven by PushT and Reacher, whereas Cube and TwoRoom remain comparatively robust to longer execution horizons.
This result highlights the importance of periodically incorporating new observations to correct accumulated execution errors.

\paragraph{Role of the training objectives.}
We ablate the three training objectives to examine their distinct roles
in RWM.
As shown in Table~\ref{tab:component-ablation},
$\mathcal L_{\mathrm{IDM}}^{\mathrm{int}}$ is critical for control:
removing it reduces PushT SR from $96.00\%$ to $39.37\%$.
In contrast, all variants containing this term achieve similar SR
($95.23$--$96.17\%$), indicating that action grounding along the constructed latent path is the primary learning signal that makes representation-space planning executable.

The other objectives provide complementary constraints on the learned representation.
$\mathcal L_{\mathrm{IDM}}^{\mathrm{enc}}$ improves physical-state
recoverability, increasing state-probe $R^2$ from $0.95$ to $0.98$,
while $\mathcal L_{\mathrm{consist}}$ geometrically aligns constructed representations with their observation-induced counterparts.
We therefore use the full objective by default, which preserves these
complementary properties while achieving SR comparable to the best
ablated variant.

\begin{table*}[htbp!]
\centering
\scriptsize
\setlength{\tabcolsep}{4.0pt}
\renewcommand{\arraystretch}{1.05}
\caption{\textbf{Component ablation of RWM.}}
    \vspace{-.2cm}
\label{tab:component-ablation}
\begin{adjustbox}{max width=\textwidth}
\begin{tabular}{lcccc}
\toprule
Loss=
& $\mathcal L_{\mathrm{IDM}}^{\mathrm{int}}$
& $\mathcal L_{\mathrm{IDM}}^{\mathrm{enc}}$
& $\mathcal L_{\mathrm{consist}}$

& PushT SR $\uparrow$\\
\midrule

$\mathcal L_{\mathrm{IDM}}^{\mathrm{int}}$ &\checkmark &&&   95.56  \\
$\mathcal L_{\mathrm{IDM}}^{\mathrm{enc}}+\mathcal L_{\mathrm{consist}}$ 
& & \checkmark &  \checkmark &   39.37 \\
$\mathcal L_{\mathrm{IDM}}^{\mathrm{int}}+\mathcal L_{\mathrm{IDM}}^{\mathrm{enc}}$ 
&  \checkmark& \checkmark &   & 96.17 \\
$\mathcal L_{\mathrm{IDM}}^{\mathrm{int}}+\mathcal L_{\mathrm{consist}}$ 
& \checkmark &  & \checkmark &  95.23\\

$\mathcal L_{\mathrm{IDM}}^{\mathrm{int}}+\mathcal L_{\mathrm{IDM}}^{\mathrm{enc}}+\mathcal L_{\mathrm{consist}}$ 
& \checkmark & \checkmark & \checkmark &{96.00} \\
\bottomrule
    \vspace{-.5cm}
\end{tabular}
\end{adjustbox}
\end{table*}

%% file: tex/conclusion.tex
\section{Conclusion}
We introduced the \emph{Representation World Model} (RWM), which learns a representation space in which executable plans can be directly constructed between states.
During training, action-grounded supervision shapes endpoint-constructed latent paths into executable plans.
At inference, RWM directly constructs and decodes such a path between the current and goal states, without recursive dynamics rollout or action-space search.
Across continuous-control benchmarks, RWM achieves strong search-free control while preserving informative state and action structure in both observed and constructed representations.
Besides, RWM also demonstrate high potential on more complex embodied control tasks.
These results suggest that representation can itself learns both high quality action-relevant states and executable plans.

%% file: tex/appendix.tex
\section{Appendix}
\subsection{Training details}
\label{app:training_setup}

We follow the data preprocessing and training protocol of
LeWM~\citep{maes2026lewm} unless otherwise specified.
All observations are resized to $224\times224$ pixels.
We apply a frame skip of 5, grouping the consecutive low-level actions
between two observations into a single action block.
We use a $90\%/10\%$ train--validation sample-level split.

\paragraph{Architecture.}
The visual encoder is a ViT-Tiny with patch size 14, trained from scratch.
Following the encoder, RWM constructs latent paths as described in Sec.~\ref{sec:method}.
The inverse dynamics model (IDM) is implemented as a three-layer MLP.
The encoder and IDM are optimized jointly, with action supervision applied
to both encoded and constructed representations.
Unlike rollout-based world models, RWM does not train an explicit forward
dynamics predictor.

\paragraph{Optimization.}
We train all models using AdamW with a learning rate of
$5\times10^{-4}$ and weight decay of $10^{-3}$ (Reacher is with $5\times10^{-5}$).
We use a linear-warmup cosine learning-rate schedule, a batch size of 128,
bfloat16 mixed precision, and gradient clipping with a maximum norm of 1.0.
Unless otherwise stated, models are trained for 10 epochs.
For the ablation studies, we use training seed 3072.
The complete training configuration is summarized in
Table~\ref{tab:training_setup}.

\begin{table}[htbp]
\centering
\caption{\textbf{Training configuration for RWM.}
We follow the LeWM training protocol wherever applicable and modify only
the model components and objectives required by RWM.}
\label{tab:training_setup}
\setlength{\tabcolsep}{7pt}
\renewcommand{\arraystretch}{1.1}
\begin{tabular}{ll}
\toprule
Setting & Value \\
\midrule
Input resolution       & $224\times224$ \\
Frame skip             & 5 \\
Train / validation split & $90\% / 10\%$ \\
Encoder network                 & ViT-Tiny/14 \\
IDM network                    & 3-layer MLP \\
IDM observation history             & 3\\
Optimizer               & AdamW \\
Learning rate           & $5\times10^{-4}$ \\
Weight decay            & $10^{-3}$ \\
LR schedule             & Linear warmup + cosine decay \\
Batch size              & 128 \\
Training epochs         & 10 \\
Precision               & bfloat16 \\
Gradient clipping       & 1.0 \\
Training seed  & 0,42,3072 \\
Training GPU & RTX$5090\times 8$\\
\bottomrule
\end{tabular}
\end{table}

\subsection{Implementation Details on Libero-Goal Test}
\label{app:libero_goal}

LIBERO-Goal differs from the continuous-control benchmarks in that the
goal observation is not available at test time. Therefore, the target
latent $z_g$ cannot be directly encoded and used to
construct a goal-conditioned latent path as in the standard RWM setting.
To handle this setting, we introduce an additional \emph{delta predictor}
that predicts the latent displacement used to construct the next representation.

During training, the full demonstration trajectory remains available.
We therefore use future states from the demonstrated trajectory to define the target latent displacements and supervise the delta predictor.
At test time, the predicted latent delta replaces the unavailable
goal-conditioned latent displacement, allows RWM to iteratively construct the latent path without access to an explicit goal representation $z_g$.

For LIBERO-Goal, we use a ViT-Small visual encoder.
The visual representation is fused with the robot proprioceptive state
before being passed to the downstream modules.
Since LIBERO contains multiple manipulation tasks with different
goal semantics, we additionally provide the task ID as conditioning
information to both the delta predictor and the inverse dynamics model
(IDM).
The delta predictor constructs the latent path, while the IDM decodes each consecutive latents into the corresponding robot action.

\subsection{Additional Physical-State Probing Visualizations}
\label{app:state_probe_visualization}

\begin{figure*}[t]
    \centering
    \includegraphics[width=\textwidth]{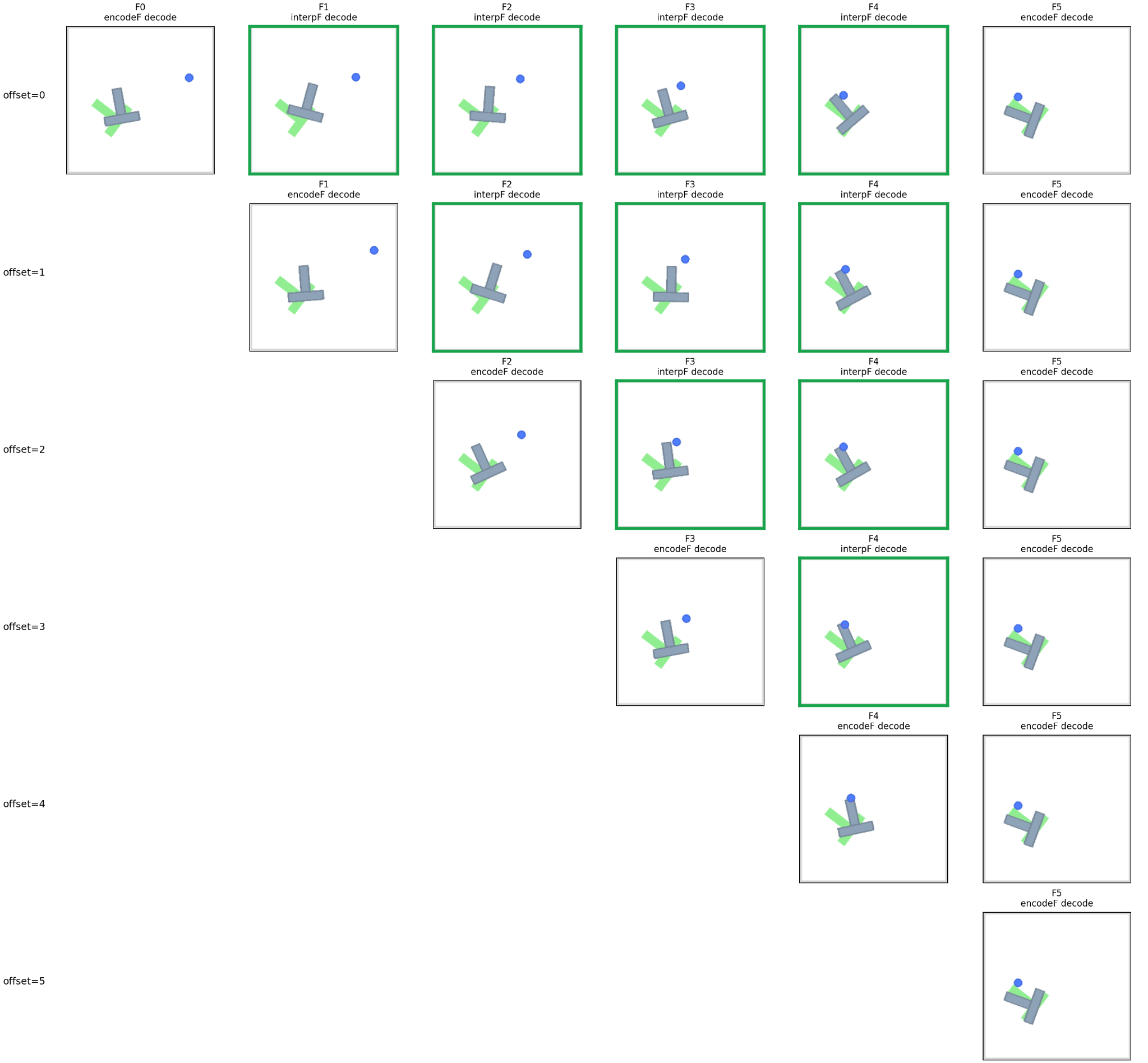}
    \caption{
    \textbf{Physical-state decoding on PushT.}
    Black-bordered images correspond to physical states decoded from encoded
    representations $z$, while green-bordered images correspond to states
    decoded from interpolated representations $\tilde{z}$.
    }
    \label{fig:state_decode_pusht}
\end{figure*}

\begin{figure*}[t]
    \centering
    \includegraphics[width=\textwidth]{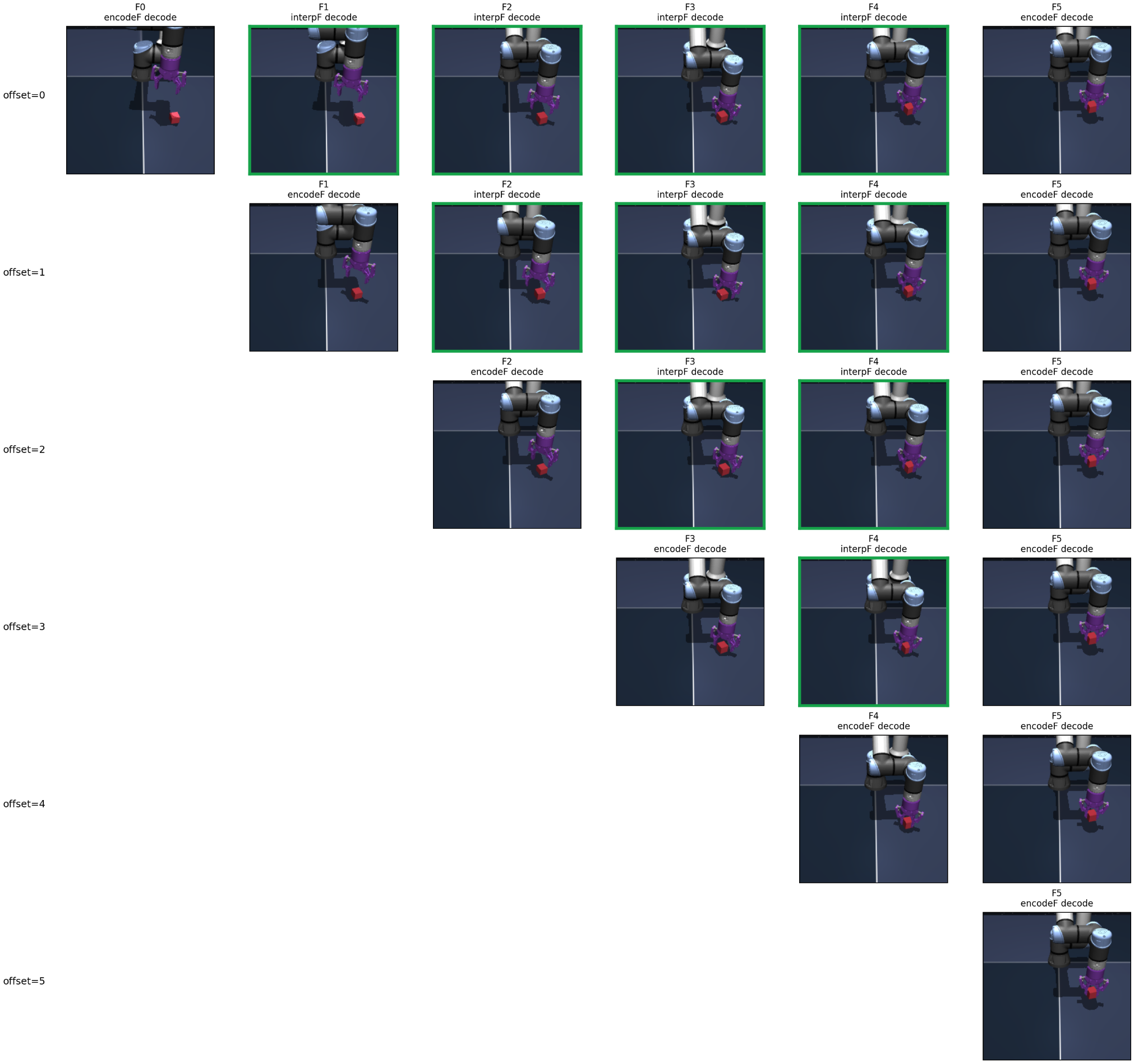}
    \caption{
    \textbf{Physical-state decoding on Cube.}
    Black-bordered images correspond to physical states decoded from encoded
    representations $z$, while green-bordered images correspond to states
    decoded from interpolated representations $\tilde{z}$.
    }
    \label{fig:state_decode_cube}
\end{figure*}

\begin{figure*}[t]
    \centering
    \includegraphics[width=\textwidth]{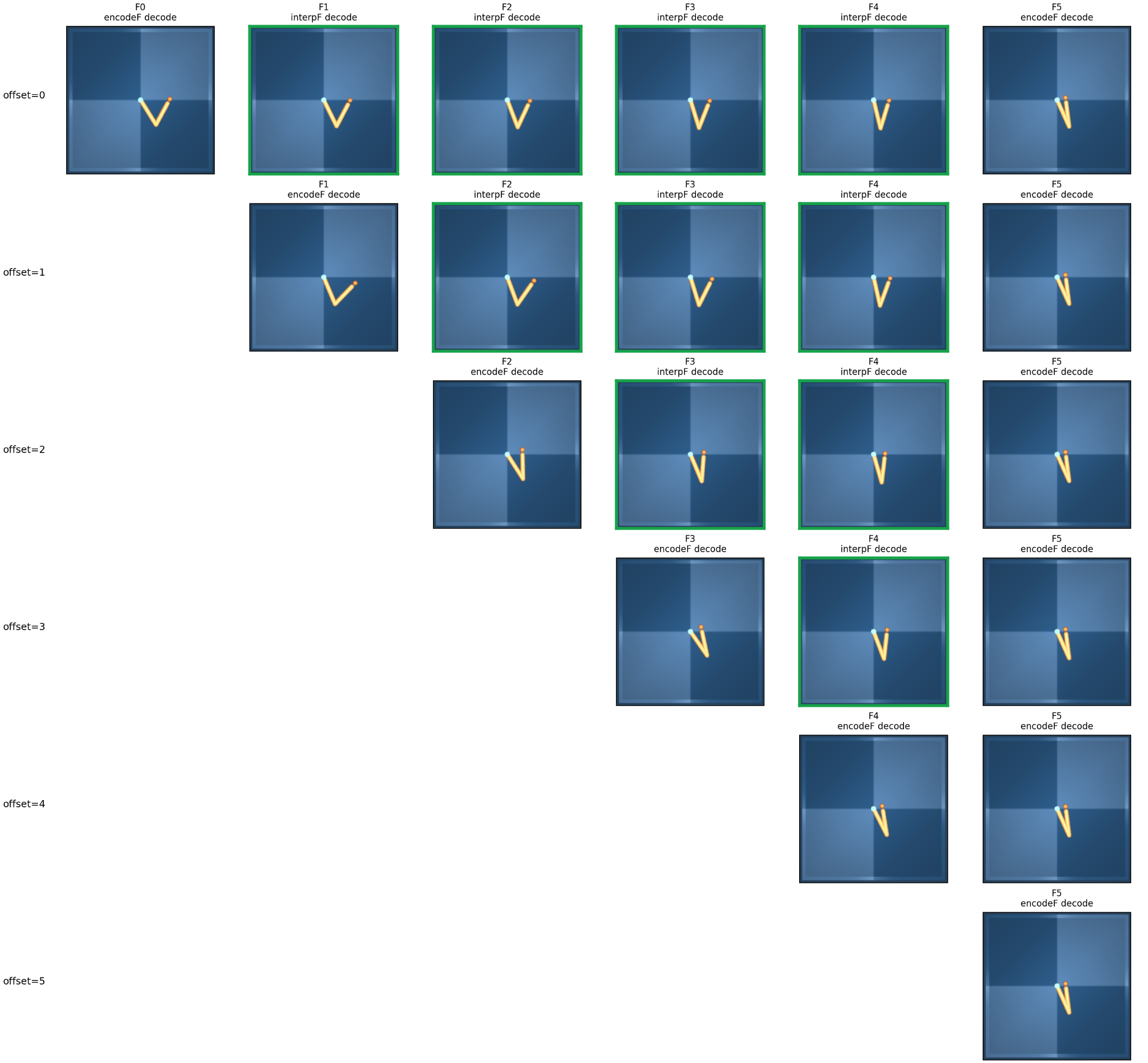}
    \caption{
    \textbf{Physical-state decoding on Reacher.}
    Black-bordered images correspond to physical states decoded from encoded
    representations $z$, while green-bordered images correspond to states
    decoded from interpolated representations $\tilde{z}$.
    }
    \label{fig:state_decode_reacher}
\end{figure*}

\begin{figure*}[t]
    \centering
    \includegraphics[width=\textwidth]{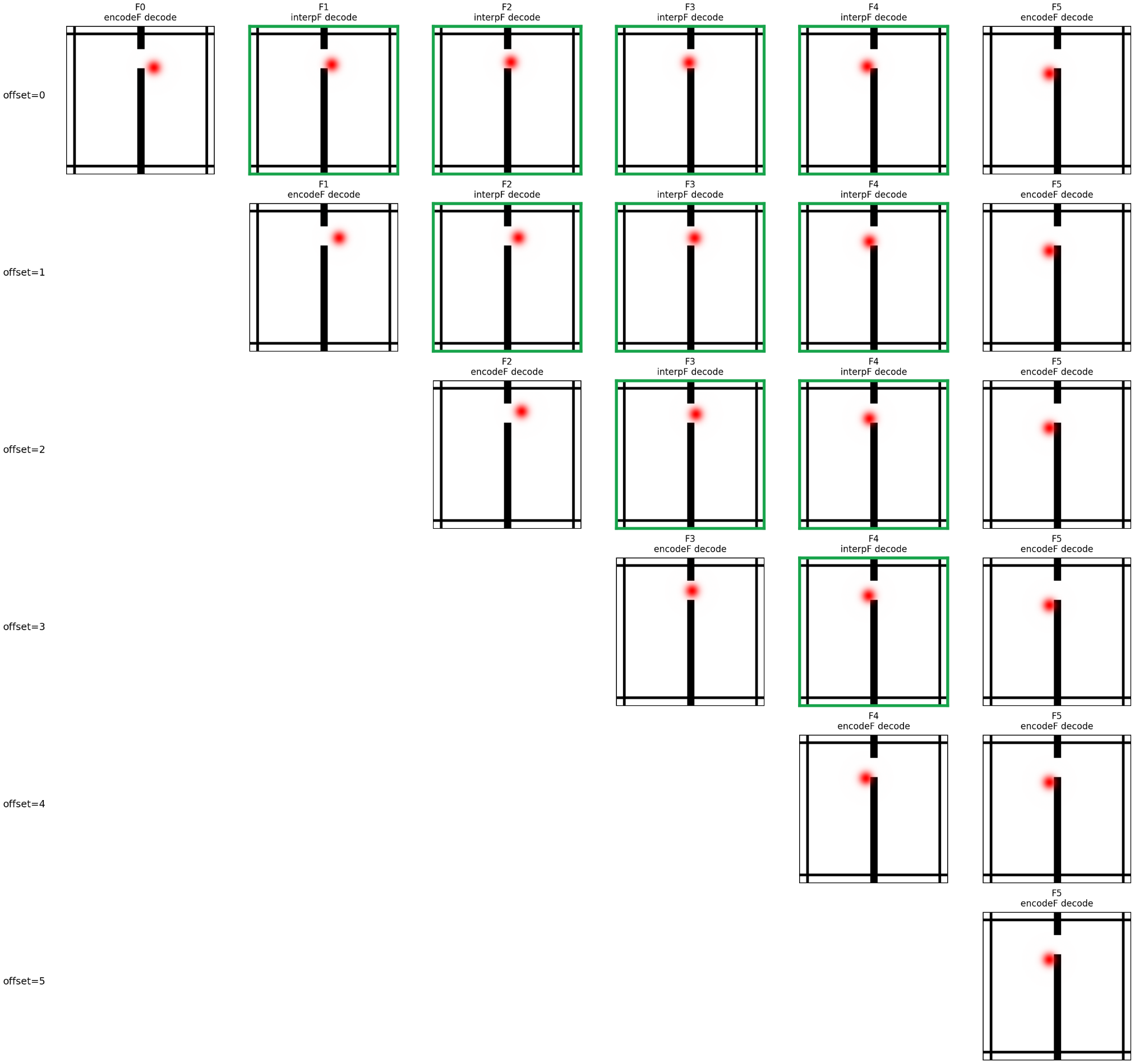}
    \caption{
    \textbf{Physical-state decoding on TwoRoom.}
    Black-bordered images correspond to physical states decoded from encoded
    representations $z$, while green-bordered images correspond to states
    decoded from interpolated representations $\tilde{z}$.
    }
    \label{fig:state_decode_tworoom}
\end{figure*}

We further provide qualitative visualizations of the physical states decoded from the learned representations in Fig.~\ref{fig:state_decode_pusht},\ref{fig:state_decode_cube},\ref{fig:state_decode_reacher},\ref{fig:state_decode_tworoom}. For each task, we apply the corresponding state probe to encoded representations $z$ and constructed interpolated representations $\tilde{z}$, and render the predicted physical states in the simulator. 
Black borders denote states decoded from encoded representations $z$, while green borders denote states decoded from interpolated representations $\tilde{z}$.
The decoded configurations from $\tilde{z}$ evolve consistently between encoded states, showing that the constructed representations preserve substantial physical-state information.

